# A SURVEY ON SENSING METHODS AND FEATURE EXTRACTION ALGORITHMS FOR SLAM PROBLEM

Adheen Ajay and D. Venkataraman

Department of Computer Science and Engineering, Amrita VishwaVidyapeetham,
Coimbatore, India
`adheen@gmail.com`
`d_venkat@cb.amrita.edu`

## ABSTRACT

*This paper is a survey work for a bigger project for designing a Visual SLAM robot to generate 3D dense map of an unknown unstructured environment. A lot of factors have to be considered while designing a SLAM robot. Sensing method of the SLAM robot should be determined by considering the kind of environment to be modelled. Similarly the type of environment determines the suitable feature extraction method. This paper goes through the sensing methods used in some recently published papers. The main objective of this survey is to conduct a comparative study among the current sensing methodsandfeature extraction algorithms and to extract out the best for our work.*

## 1. INTRODUCTION

It is in the beginning of twentieth century, the world first introduced with Robots. As the time passed, robotic field has grown up and achieved peaks within one or two decades dramatically. Highly accurate and specific robots are widely used in many applications now, including medical field, construction field and even in disaster management situations. Even though the robotic field has achieved tremendous progress,modelling of environments using SLAM is still being a challenging problem. SLAM is Simultaneous Localization and Mapping. It is also called as Concurrent Mapping and Localization (CML). The basic objective of SLAM problem is to generate a map of an environment using a mobile robot. Such maps have applications in robot navigation, manipulation,tele-presence, semantic mappingand unmanned vehicles and also in planet rovers. This survey is conducted as the initialization of a bigger project of designing a Visual SLAM robot to generate 3D dense map of an unknown unstructured static indoor environment.

The paper is organized as follows. Introduction is given in section 1. Section 2 gives an idea about SLAM. Survey on sensing methods is carried out in section 3. Survey on feature extraction algorithms is done in section 4. Comparison of the sensing methods and feature extraction algorithms are done and conclusions are given in section 5.

## 2. SLAM PROBLEM

SLAM is one of the most widely researched subfields of robotics. The scenario behind a SLAM problem is explained here. Consider a simple mobile robot: a set of wheels connected to motors and a controlling processor.A camera is also there as inputting device. Now imagine the robot being used remotely by an operator to map inaccessible places. The robot moves around the environment, and the camera provides enough visual information for the operator to understand where surrounding objects are and how the robot is oriented in reference to them. What the human operator is doing is an example of SLAM (Simultaneous Localization and Mapping). Determining the location of objects in the environment is a case of mapping, and establishing the robot position with respect to these objects is an example of localization. The SLAM subfield of robotics attempts to provide a way for robots to do SLAM autonomously. A solution to the SLAM problem would allow a robot to make maps without any human assistance.This paper focuses on building a 3D dense map of the environment. The robot is equipped with a sensor and a processor, it moves through an environment and finally comes out with a single 3D map of the environment. We can say that a SLAM process is an active process where it updates the generated 3D map when a new inputs areoccuring. The overall system diagram of the work is given in Figure 1.

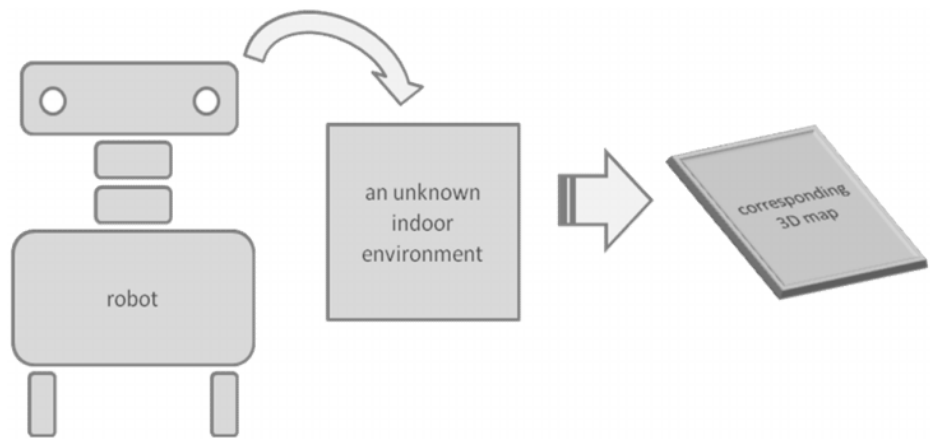

Figure 1. Overall system design

The type of environment that the robot meant to use is a factor in determining the best suitable sensing methods and feature extraction methods. The environment for our work is an indoor environment. It will not be properly structured and no information about the environment will be preloaded on the robot. So we can summarize that the environment is an unknown unstructured indoor environment. This survey will help to find the best suitable sensing methods and feature extraction method for the current problem.

## 3. SURVEY ON SENSING METHODS

Sensors for SLAM are generally classified into three based on sensing methodology. They are sonar based sensors, laser based sensors and vision based sensors.These types have sensors for getting 2D and 3D data. We will be concentrating only on 3D sensors as our aim is on designing a system for 3D modelling.Laser range finders are used as 3D sensors in [1], where the robot can detect a three dimensional range relative to the global horizon plane. It uses slope sensors also in order to amend the data but its implementation is quite costly and difficult. It gives only the depth information but no color information of the scene.Weizhen Zhou [2] presented a 3D SLAM by using a conventional RGB camera and a time-of-flight range camera. The range camera uses IR projection and capturing. The range camera is of low resolution so the 3D information obtained will not be clear and accurate. Another challenge in this work is that the two cameras should be in proper alignment to have accurate 3D information. In the literature by Daniele Marzorati[3], sensor is a trinocular camera system. Trinocular camera is an arrangement of three RGB cameras to capture same scene. Search for the corresponding pixels in a stereo camera is speeded up in trinocular camera. It is more accurate than a stereo camera. In the work of Lina M. Paz [4], 3D sensing is done using a stereo vision camera- bumble bee. It has lower resolution and it purely depends upon the illumination on the scene. It also assumes pixel correspondence between the stereo frames. PeterHenry[5] uses Kinect to generate a 3D dense map. Kinect is highly accurate and precise. It give accurate 3D point cloud. It provides high resolution depth video($640 \times 480$). Kinect uses IR projection to get the depth information.Compared to Bumblebee camera, Kinect is cheaper too. Kinect is released as a motion sensor device for the gaming console Microsoft Xbox 360. Later its possibilities are exploited inresearch applications.Different sensing methods and its observations are tabulated in Table 1.

Table 1. Observations on different sensing methods.

| Ref | Sensing Method | Observations |
| --- | --- | --- |
| [1] | Laser range finder and slope sensors. | • The robot can detect a three dimensionalrange data relative to the global horizontalplane.<br>• Relatively difficult to implement it. |
| [2] | A time-of-flight range camera (Swiss Ranger SR-3000, $176 \times 144$ pixels) A conventional camera (Point Grey Dragonfly2, $1024 \times 768$ pixels) | • Two cameras should be aligned properly to capture the same scene<br>• The range sensor has low resolution<br>• Swiss Ranger applies time-of-flight concepts using IR ray projection and perception |
| [3] | Trinocular stereo system. | • Uses 3 RGB cameras<br>• Speeds up the search for triplets corresponding 2D segments in the 3 images<br>• More accurate<br>• Alignment problem |
| [4] | Stereo Camera- Bumble Bee | • Produces dense 3D maps.<br>• Limited accuracy<br>• High cost |
| [5] | Microsoft Xbox360 Kinect | • Highly accurate and precise 3D information<br>• Resolution is high ($640 \times 480$)<br>• Uses IR lasers to get the depth information<br>• Much faster performance |

## 4. SURVEY ON FEATURE EXTRACTION METHODS

Feature based SLAM robots make use of feature points in the scene video to track the relative motion of the robot in the environment. Different feature extraction methods can be used to extract features for a SLAM problem. The main objective of any feature extraction problem is to get features with maximum information. The suitable features detection algorithm will be different for different environments. Here our goal is to find the best suitable feature detection algorithm for this work. Work in [3] uses Harris corner detection, which is faster in performance. Lina M. Paz [4] uses Shi-Thomasi feature tracking algorithm to find the feature points in the image. These features are robust than Harris corners and are more suitable for tracking. Weizhen Zhou[2] prefers Scale Invariant Feature Tracking(SIFT) for SLAM problems. SIFT is more robust to noise and scale variations. Observations are given in Table 2.

| Ref | Contents | Observations |
|-----|----------|--------------|
| [3] | Harris Corner Detection | • More location accuracy<br>• Relatively faster |
| [4] | Shi Thomasi Corner Detection | • More robust than Harris Corner detection<br>• Suitable for tracking<br>• Faster execution |
| [2] | SIFT – Scale Invariant Feature Transform | • Blob detector<br>• Robust features<br>• Low performance at the corners<br>• Suitable for tracking |

## 5. DISCUSSION

From the above observations we need to reach in a conclusion. Laser range finders were common a decade back, but none is using that because of its low accuracy and high implementation difficulty in indoor applications. Time of flight camera, Swiss ranger is a good option for our work, but its low resolution and low accuracy are still its drawbacks. Trinocular cameras gives accurate 3D information. But proper alignment have to be maintained among the cameras, also it suffers from brightness constancy problem. Stereo cameras are widely using sensor for getting dense 3D information. It uses two similar cameras to capture the same scene, with a small inter camera distance. But accuracy of this camera depends on the illumination and it definitely suffers from brightness constancy problem. Bumble bee camera is an example for stereo camera. Kinect is the latest trend in 3D scene capture for small ranges. It uses a RGB camera and an IR depth camera together and combines the output to get the 3D point cloud of the scene. It gives highly accurate dense 3D point cloud in the range of 1 to 10 meters from it. It is cheaper than bumblebee camera and doesn't suffer from brightness constancy problem. Since the sensor's highly productive range is comparable with the indoor dimensions we can conclude that Kinect is more suitable 3D sensor for our work.

Harris corner detection, Shi-Thomasi feature detector and SIFT are the common feature extraction algorithms in SLAM. Harris corner detector and Shi Thomasi corner detectors extracts the most informative points in the scene- the corners. These features are more effective in structured environments, or in environments where there is enough corner points. In an unstructured environment we cannot expect a productive number of such feature points. SIFT feature can be effective in such environments. SIFT is a blob detector,treats blobs in a scene as features than corners. So SIFT can be used as feature detectors for our work.

## AUTHORS

**Adheen Ajay** was born in Kerala, India, in 1989. He received the B.Tech degree in Electronics and Communication from the University of Calicut, Kerala in 2010. He has been working towards the M.Tech degree in Computer Vision and Image Processing under the Department of Computer Science, Amrita VishwaVidyapeetham, Coimbatore, India. His current research interests include 3D modelling of environments using computer vision and Simultaneous Localisation and Mapping (SLAM).

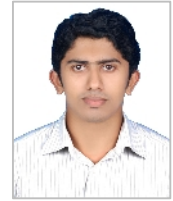